\documentclass[10pt,twocolumn,letterpaper]{article}

\usepackage{wacv}
\makeatletter
\@namedef{ver@everyshi.sty}{}
\makeatother
\usepackage{tikz}
\usepackage{times}
\usepackage{epsfig}
\usepackage{graphicx}
\usepackage{amsmath}
\usepackage{amssymb}
\usepackage{booktabs}
\usepackage[accsupp]{axessibility}  
\usepackage{stfloats}
\usepackage{bm}
\usepackage{flushend}
\usepackage{lipsum}
\usepackage[numbers, sort, compress]{natbib}
\usepackage[ruled,vlined]{algorithm2e}
\SetAlFnt{\small}
\SetAlCapFnt{\small}
\usepackage{enumitem}
\usepackage{arydshln}
\usepackage{colortbl}
\usepackage{tabularx}
\usepackage{ragged2e}
\usepackage{pict2e}
\usepackage{array}
\usepackage{threeparttablex}
\usepackage{makecell}
\usepackage{overpic}
\usepackage{pifont}
\let\oldding\ding
\renewcommand{\ding}[2][1]{\scalebox{#1}{\oldding{#2}}}
\newcommand{\cmark}{\ding{51}}%
\newcommand{\xmark}{\ding{55}}%
\usepackage{color}
\usepackage{url}
\usepackage{bm}
\usepackage[scr=boondoxo]{mathalpha}
\usepackage{mathtools, nccmath}

\usepackage{adjustbox}
\usepackage{setspace}
\usepackage{xargs}

\usepackage{caption}
\usepackage{subcaption}

\usepackage[utf8]{inputenc} 
\usepackage[T1]{fontenc}    
\usepackage{url}            
\usepackage{amsfonts}       
\usepackage{xfrac}          
\usepackage{microtype}      
\usepackage{csquotes}

\usepackage{comment}

\newcolumntype{Y}{>{\centering\arraybackslash}X}
\makeatletter
\def\adl@drawiv#1#2#3{%
        \hskip.5\tabcolsep
        \xleaders#3{#2.5\@tempdimb #1{1}#2.5\@tempdimb}%
                #2\z@ plus1fil minus1fil\relax
        \hskip.5\tabcolsep}
\newcommand{\cdashlinelr}[1]{%
  \noalign{\vskip\aboverulesep
           \global\let\@dashdrawstore\adl@draw
           \global\let\adl@draw\adl@drawiv}
  \cdashline{#1}
  \noalign{\global\let\adl@draw\@dashdrawstore
           \vskip\belowrulesep}}
\makeatother

\newcommand{\sampclr}{\texttt{SAMP-CLR}}
\newcommand{\samp}{\texttt{SAMP}}
\newcommand{\opttune}{\texttt{OpT-Tune}}
\newcommand{\ourmethod}{\texttt{SAMPTransfer}}
\newcommand{\miniImagenet}{\textit{mini}ImageNet}
\newcommand{\tieredImagenet}{\textit{tiered}ImageNet}
\newcommand{\cfs}{\texttt{Conv$4$}}
\newcommand{\cfb}{\texttt{Conv$4$b}}

\def\wacvPaperID{1457} 

\wacvalgorithmstrack   

\wacvfinalcopy 

\ifwacvfinal
\usepackage[breaklinks=true,bookmarks=false, colorlinks]{hyperref}
\else
\usepackage[pagebackref=true,breaklinks=true,colorlinks,bookmarks=false]{hyperref}
\fi
\usepackage{doi}
\usepackage[nameinlink, capitalize]{cleveref}

\begin{document}

\title{Self-Attention Message Passing for Contrastive Few-Shot Learning}


\author{Ojas Kishorkumar Shirekar$^{1,2}$, Anuj Singh$^{1,2}$, Hadi Jamali-Rad$^{1,2}$\\
$^1$Delft University of Technology (TU Delft), The Netherlands  \\
$^2$Shell Global Solutions International B.V., Amsterdam, The Netherlands \\
{\tt\small \{o.k.shirekar, a.r.singh\}@student.tudelft.nl, h.jamalirad@tudelft.nl}
}

\maketitle
\thispagestyle{empty}

\vspace{-0.3cm}
\begin{abstract}
\vspace{-0.3cm}

Humans have a unique ability to learn new representations from just a handful of examples with little to no supervision. Deep learning models, however, require an abundance of data and supervision to perform at a satisfactory level. Unsupervised few-shot learning (U-FSL) is the pursuit of bridging this gap between machines and humans. Inspired by the capacity of graph neural networks (GNNs) in discovering complex inter-sample relationships, we propose a novel self-attention based message passing contrastive learning approach (coined as \sampclr) for U-FSL pre-training. We also propose an optimal transport (OT) based fine-tuning strategy (we call \opttune) to efficiently induce task awareness into our novel end-to-end unsupervised few-shot classification framework (\ourmethod). Our extensive experimental results corroborate the efficacy of \ourmethod{} in a variety of downstream few-shot classification scenarios, setting a new state-of-the-art for U-FSL on both \miniImagenet{} and \tieredImagenet{} benchmarks, offering up to $7\%+$ and $5\%+$ improvements, respectively. Our further investigations also confirm that \ourmethod{} remains on-par with some supervised baselines on \miniImagenet{} and outperforms all existing U-FSL baselines in a challenging cross-domain scenario. Our code can be found in our GitHub repository: \url{https://github.com/ojss/SAMPTransfer/}.\footnote{This paper is accepted to appear in the proceedings of WACV 2023.}

\end{abstract}








\vspace{-0.4cm}
\section{Introduction}
\label{sec:intro}
\vspace{-0.2cm}
Deep learning models have become increasingly large and data hungry to be able to guarantee acceptable downstream performance. Humans need neither a ton of data samples nor extensive forms of supervision to understand their surroundings and the semantics therein. Few-shot learning has garnered an upsurge of interest recently as it underscores this fundamental gap between humans' adaptive learning capacity compared to data-demanding deep learning methods. In this realm, few-shot classification is cast as the task of predicting class labels for a set of unlabeled data points (\textit{query set}) given only a small set of labeled ones (\textit{support set}). Typically, query and support data samples are drawn from the same distribution. 

Few-shot classification methods usually consist of two sequential phases: (i) \textit{pre-training} on a large dataset of \textit{base} classes, regardless of this pre-training being supervised or unsupervised. This is followed by (ii) \textit{fine-tuning} on an unseen dataset consisting of \textit{novel} classes. Normally, the classes used in the pre-training and fine-tuning are mutually exclusive. In this paper, we focus on the self-supervised setting (also interchangeably called \textquote{unsupervised} in literature) where we have no access to the actual class labels of the \textquote{base} dataset. Our motivation to tackle unsupervised few-shot learning (U-FSL) is that it poses a more realistic challenge, closer to humans' learning process. 

The body of work around U-FSL can be broadly classified into two different approaches. The first approach relies on the use of \textit{meta-learning} and episodic pre-training that involves the creation of synthetic \textquote{tasks} to mimic the subsequent episodic fine-tuning phase \citep{Finn2017Model-agnosticNetworks, Hsu2018UnsupervisedMeta-Learning, Khodadadeh2018UnsupervisedClassification, Antoniou2019AssumeAugmentation, Ye2022, lee2021meta, Ji2019UnsupervisedTraining}. 
The second approach follows a \textit{transfer learning} strategy, where the network is trained non-episodically to learn optimal representations in the pre-training phase from the abundance of unlabeled data and is then followed by an episodic fine-tuning phase \citep{Medina2020Self-SupervisedClassification, shirekar2022self, dhillon2019baseline}. To be more specific, a feature extractor is first pre-trained to capture the structure of unlabeled data (present in base classes) using some form of representation learning \citep{Medina2020Self-SupervisedClassification, chen2020simple, caron2020unsupervised, shirekar2022self}.
Next, a prediction layer (a linear layer, by convention) is fine-tuned in conjunction with the pre-trained feature extractor for a swift adaptation to the novel classes.
The better the feature extractor models the distribution of the unlabeled data, the less the predictor requires training samples, and the faster it adapts itself to the unseen classes in the fine-tuning and eventual testing phases. Some recent studies \citep{Medina2020Self-SupervisedClassification, goodemballneed2020, das2022confess} argue that transfer learning approaches outperform meta-learning counterparts in standard in-domain and cross-domain settings, where base and novel classes come from totally different distributions. 

On the other side of the aisle, supervised FSL approaches that follow the episodic training paradigm may include a certain degree of \textit{task awareness}.
Such approaches exploit the information available in the query set during the training and testing phases \citep{xu2021attentional, ye2020few, Cui2021} to alleviate the model's sample bias. As a result, the model learns to generate task-specific embeddings by better aligning the features of the support and query samples for optimal metric based label assignment.
Some other supervised approaches do not rely purely on convolutional feature extractors. Instead, they use graph neural networks (GNN) to model instance-level and class-level relationships \citep{garcia2018fewshot, kim2019edge, yu2022hybrid, yang2020dpgn}. This is owing to the fact that GNN's are capable of exploiting the manifold structure of the novel classes \citep{xiaojin2002learning}. However, looking at the recent literature, one can barely see any GNN based architectures being used in the unsupervised setting.

Recent unsupervised methods use a successful form of \textit{contrastive learning} \citep{chen2020simple} in their self-supervised pre-training phase. Contrastive learning methods typically treat each image in a batch as its own class. The only other images that share the class are the augmentations of the image in question. 
Such methods enforce similarity of representations between pairs of an image and its augmentations (positive pairs), while enforcing dissimilarity between all other pairs of images (negative pairs) through a \textit{contrastive loss}.
Although these methods work well, they overlook the possibility that within a randomly sampled batch of images there could be several images (apart from their augmentations) that in reality belong to the same class. By applying the contrastive loss, the network may inadvertently learn different representations for such images and classes.

To address this problem, recent methods such as SimCLR \citep{chen2020simple} introduce larger batch sizes in the pre-training phase to maximize the number of negative samples.
However, this approach faces two shortcomings: (i) increasingly larger batch sizes, mandate more costly training infrastructure, and (ii) it still does not ingrain intra-class dependencies into the network. Point (ii) still applies to even more recent approaches, such as ProtoCLR \citep{Medina2020Self-SupervisedClassification}. A simple yet effective remedy of this problem proposed in C$^3$LR \citep{shirekar2022self} where an intermediate clustering and re-ranking step is introduced, and the contrastive loss is accordingly adjusted to ingest a semblance of class-cognizance. However, the problem could be approached from a different perspective, where the network explores the structure of data samples per batch.

We propose a novel U-FSL approach (coined as \ourmethod{}) that marries the potential of GNNs in learning the global structure of data in the pre-training stage, and the efficiency of optimal transport (OT) for inducing task-awareness in the following fine-tuning phase.  More concretely, with \ourmethod{} we introduce a novel self-attention message passing contrastive learning (\sampclr) scheme that uses a form of \textit{graph attention} allowing the network to learn refined representations by looking beyond single-image instances per batch. Furthermore, the proposed OT based fine-tuning strategy (we call \opttune) aligns the distributions of the support and query samples to improve downstream adaptability of the pre-trained encoder, without requiring any additional parameters. Our contributions can be summarized as: 
\begin{enumerate}
    \item We propose \ourmethod{}, a novel U-FSL approach that introduces a self-attention message passing contrastive learning (\sampclr) paradigm for unsupervised few-shot pre-training.
    \item We propose applying an optimal transport (OT) based fine-tuning (\opttune) strategy to efficiently induce task-awareness in both fine-tuning and inference stages.
    \item We present a theoretical foundation for \ourmethod{}, as well as extensive experimental results corroborating the efficacy of \ourmethod{}, and setting a new state-of-the-art (to our best knowledge) in both \miniImagenet{} and \tieredImagenet{} benchmarks, we also report competitive performance on the challenging CDFSL benchmark  \cite{guo2019new}.
\end{enumerate}

\section{Related Work}
\vspace{-0.1cm}
\label{sec:rel-work}
\textbf{Self-Supervised learning}. Self-supervised learning (SSL) is a term used for a collection of unsupervised methods that obtain supervisory signals from within the data itself, typically by leveraging the underlying structure in the data. 
The general technique of self-supervised learning is to predict any unobserved (or property) of the input from any observed part.
Several recent advances in the SSL space have made waves by eclipsing their fully supervised counterparts \citep{Goyal2021}. Some examples of seminal works include SimCLR \citep{chen2020simple}, BYOL \citep{grill2020bootstrap}, SWaV \citep{caron2020unsupervised}, MoCo \citep{he2020momentum}, and SimSiam \citep{chen2021exploring}. Our pre-training method $\sampclr$ is inspired by SimCLR \citep{chen2020simple}, ProtoTransfer \citep{Medina2020Self-SupervisedClassification} and C$^3$LR \citep{shirekar2022self}.

\textbf{Metric learning}. Metric learning aims to learn a representation function that maps the data to an embedding space. The distance between objects in the embedding space must preserve their similarity (or dissimilarity) - similar objects are closer, while dissimilar objects are farther. For example, unsupervised methods based on some form of contrastive loss, such as SimCLR \citep{chen2020simple} or NNCLR \citep{dwibedi2021little}, guide objects belonging to the same potential class to be mapped to the same point and those from different classes to be mapped to different points. Note that in an unsupervised setting, each image in a batch is its own class. This process generally involves taking two crops of the same image and encouraging the network to emit an identical representation for the two, while ensuring that the representations remain different from all other images in a given batch. 
Metric learning methods have been shown to work quite well for few-shot learning. AAL-ProtoNets \citep{Antoniou2019AssumeAugmentation}, ProtoTransfer \citep{Medina2020Self-SupervisedClassification}, UMTRA \citep{Khodadadeh2018UnsupervisedClassification}, and certain GNN methods \citep{garcia2018fewshot} are excellent examples that use metric learning for few-shot learning.

\textbf{Graph Neural Networks for FSL}. Since the first use of graphs for FSL in \citep{garcia2018fewshot}, there have been several advancements and continued interest in using graphs for supervised FSL. 
In \citep{garcia2018fewshot}, each node corresponds to one instance (labeled or unlabeled) and is represented as the concatenation of a feature embedding and a label embedding. The final layer of their model is a linear classifier layer that directly outputs the prediction scores for each unlabeled node.
There has also been an increase in methods that use transduction. TPN \citep{liu2018learning} is one of those methods that uses graphs to propagate labels \citep{xiaojin2002learning} from labeled samples to unlabeled samples. Although methods such as EGNN \citep{kim2019edge} make use of edge and node features, earlier methods focused only on using node features.
Graphs are attractive, as they can model intra-batch relations and can be extended for transduction, as used in \citep{kim2019edge, liu2018learning}. In addition to transduction and relation modeling, graphs are highly potent as task adaptation modules. HGNN \citep{yu2022hybrid} is an example in which a graph is used to refine and adapt feature embeddings. It must be noted that most graph-based methods have been applied in the supervised FSL setting. To the best of our knowledge, we are the first
to use it in any form for U-FSL. More specifically, we use a message passing network as a part of our network architecture and pre-training scheme.

\begin{figure*}[th!]
    \centering
    \begin{overpic}[abs, unit=1cm, width=0.88\textwidth, trim=0.5cm 1.6cm 0.5cm 0.5cm, clip, percent]{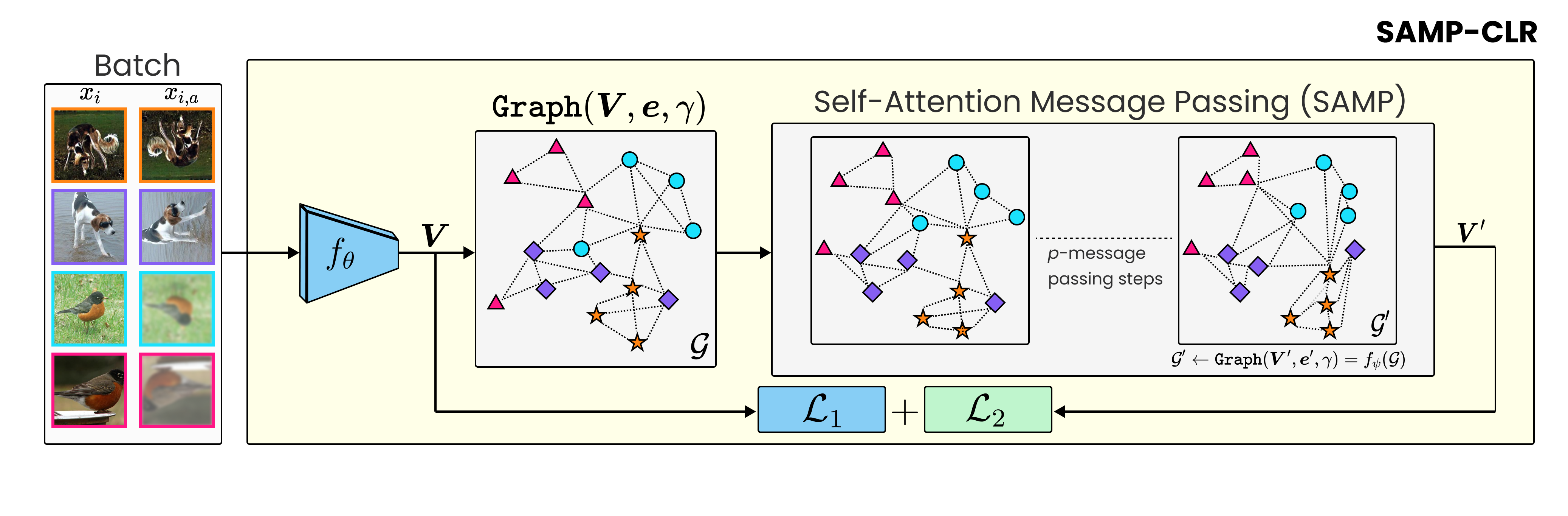}
    \put(16, 23) {\footnotesize\(f_\Omega=f_\theta \circ f_\psi\)}
    \put(0,0){\color{cyan}\linethickness{0.2mm}
              \polygon(15.5, 22)(29, 22)(29, 25)(15.5, 25)}
    \end{overpic}
    \vspace{-0.1cm}
    \caption{\sampclr\ schematic view and pre-training procedure. In the figure, $x_{i, a}$ is an image sampled from the augmented set $\mathcal{A}$. The $p$-message passing steps refine the features extracted using a CNN encoder.}
    
    \label{fig:pre-training}
\end{figure*}

\begin{figure*}[t]
    \centering
    \begin{overpic}[abs, unit=1cm, width=0.8\textwidth, trim=0.5cm 0.2cm 0.2cm 0.75cm, clip, percent]{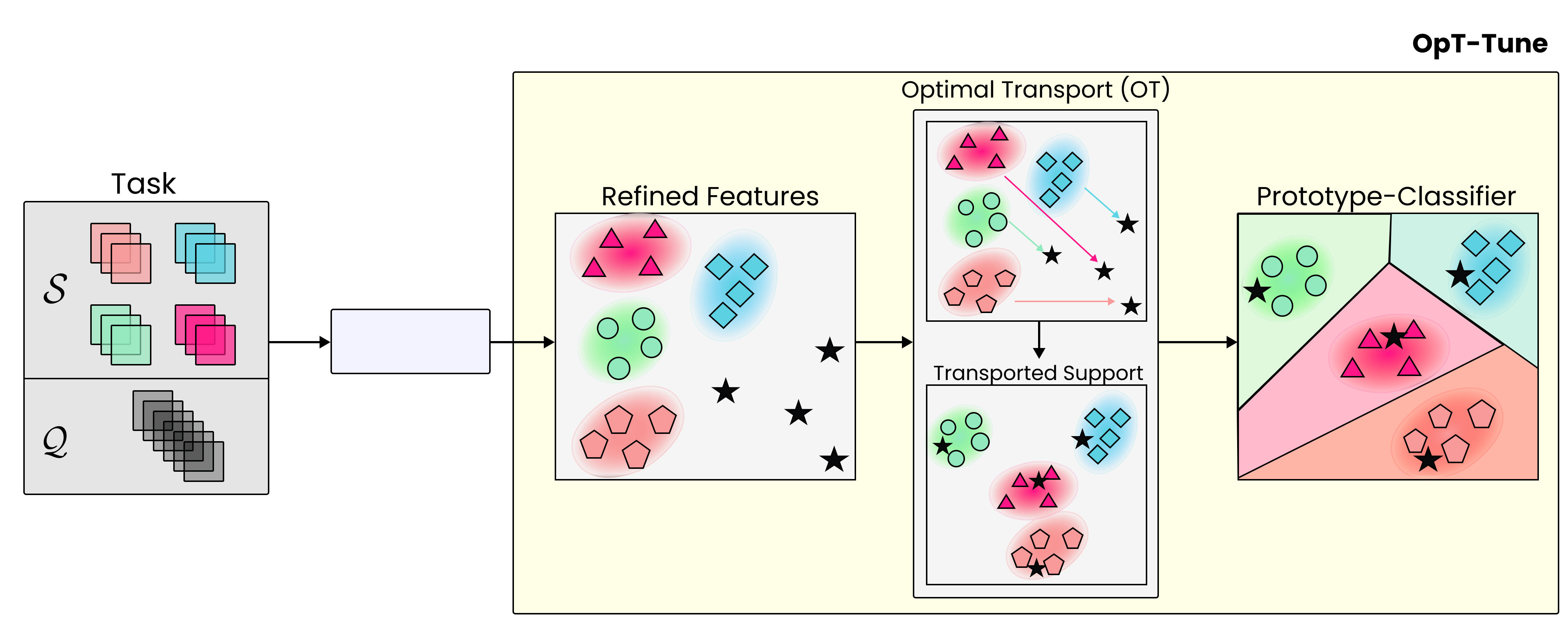}
    \put(23.6, 17.6){\large \(f_\Omega\)}
    \end{overpic}
    \vspace{-0.1cm}
    \caption{Features extracted from the pre-trained CNN are used to build a graph. The features are first refined using the pre-trained \samp{} layer(s). Then \opttune{} aligns support features with query features.}
    
    \label{fig:classification}
    \vspace{-0.3cm}
\end{figure*}
\section{Proposed Method (\ourmethod)}
\label{sec:proposed-method}

In this section, we first describe our problem formulation. We then discuss the two subsequent phases of the proposed approach: (i) self-supervised pre-training ($\sampclr$), and (ii) the optimal transport based episodic supervised fine-tuning ($\opttune$). Together, these two phases constitute our overall approach, which we have coined as \ourmethod. The mechanics of the proposed pre-training and fine-tuning procedures are illustrated in \cref{fig:pre-training,fig:classification}, respectively.
\subsection{Preliminaries}
\label{ssec:preliminaries}
Let us denote the training data of size $D$ as $\mathcal{D}_{\textup{tr}} = \{({\bm x}_i, y_i)\}_{i = 1}^{D}$ with $({\bm x}_i, y_i)$ representing an image and its class label, respectively. In the pre-training phase, we sample $L$ random images from $\mathcal{D}_{\textup{tr}}$ and augment each sample $A$ times by randomly sampling augmentation functions $\zeta_a(.), \forall a \in [A]$ from the set $\mathcal{A}$. This results in a mini-batch of size $B = (A + 1)L$ total samples. Note that in the unsupervised setting, we have no access to the data labels in the pre-training phase. Next, we fine-tune our model episodically \cite{Vinyals2016MatchingLearning} on a set of randomly sampled tasks $\mathcal{T}_i$ drawn from the test dataset $\mathcal{D}_{\textup{tst}} = \{({\bm x}_i, y_i)\}_{i = 1}^{D^\prime}$ of size $D^\prime$. A task, $\mathcal{T}_i$, is comprised of two parts: (i) the support set $\mathcal{S}$ from which the model learns, (ii) the query set $\mathcal{Q}$ on which the model is evaluated. The support set $\mathcal{S}=\{\bm{x}^{s}_i, y^{s}_i\}_{i=1}^{NK}$ is constructed by drawing $K$ labeled random samples from $N$ different classes, resulting in the so-called ($N$-way, $K$-shot) settings.
The query set $\mathcal{Q}=\{\bm{x}^{q}_j\}_{j=1}^{NQ}$ then contains $NQ$ unlabeled samples. By convention, we denote $\mathcal{T}_i = \mathcal{S}_i \cup \mathcal{Q}_i$ by $(N, K)$.  
\subsection{Self-Attention Message Passing (\samp{})}\label{ssec:samp-explainer}
Our network architecture consists of a convolutional (CNN) feature extractor $f_{\theta}$ and a message passing network based on self-attention, $f_{\psi}$.
The CNN feature extractor $f_\theta$, parameterized by $\theta$, is used to extract features $\bm{V} = f_\theta(\bm{X})$, where $\bm{V} \in \mathbb{R}^{B \times d}$ is the set of $B$ features each of size $d$ and $\bm{X} \in \mathbb{R}^{B \times C\times H \times W}$ is a batch of $B$ images of size $C\times H \times W$. To help refine the features and use batch-level relationships, we create a graph $\mathcal{G} = \texttt{Graph}(\bm{V}, \bm{e}, \gamma)$ where $\bm V$ is treated as a set of initial node features, $\bm{e}$ is the pairwise distance between all nodes based on a given distance metric and $\gamma$ is a threshold on the values in $\bm e$ that determines whether two nodes will be connected or not. Note that $|\mathcal{G}| = B$, as we build the graph over the $B$ samples in our batch. We use a \textbf{s}elf-\textbf{a}ttention \textbf{m}essage \textbf{p}assing neural network (we call \samp{}) $f_\psi$, parameterized by $\psi$, to refine the initial feature vectors by exchanging and amalgamating information between all pairs of connected nodes. 
From now on, as can be seen in \cref{fig:pre-training,fig:classification}, we refer to the combination of the feature extractor $f_\theta$ and the \samp{} module $f_\psi$ as $f_\Omega = f_\psi \circ f_\theta$ where $\Omega = \{\theta, \psi\}$ is the collection of all parameters. The \samp{} layers, $f_\psi$ operate on the graph $\mathcal{G}$.

To allow an effective exchange of information to refine initial node features $\bm{V}$, we make use of graph attention in a slightly different manner than the standard graph attention defined in \citep{velic018graph}.
The graph attention in \citep{velic018graph} uses a single weight matrix $\bm{W}$ that acts as a shared linear transformation for all nodes. Instead, we choose to use scaled dot-product self-attention as defined in \citep{vaswani2017attention, seidenschwarz2021learning}.
The major benefit of this design choice is that it enhances the network with more expressivity, as shown in \citep{brody2021attentive, kim2022pure}.
Notably, the use of three separate representations (query, key, and value) instead of just a single weight matrix to linearly transform the data is key to modeling relationships between data points. 

We apply $p$ successive message passing steps similar to \citep{velic018graph, seidenschwarz2021learning}. In each step, we pass messages between the connected nodes of $\mathcal{G}$ and obtain updated features in ${\bm V}^{p+1}$, at step $p+1$. 
Here, the $i$-th row of ${\bm V}^{p+1}$ is given by $\bm{V}_{i}^{p+1}=\sum_{j \in \mathcal{N}_{i}} \lambda_{i j}^{p} \bm{W}^{p} \bm{V}_{j}^{p}$, where $\lambda_{i j}$ is the attention score between the nodes $i$ and $j$, $\bm{W}^p \in \mathbb{R}^{d \times d}$ is the message passing weight matrix at step $p$, and $\mathcal{N}_i$ denotes the set of neighboring nodes of node $i$. In this way, $\lambda_{i,j}$ allows our update mechanism to flexibly weight every sample w.r.t every other sample in the batch. We employ scaled dot-product self-attention to compute attention scores, leading to: $\lambda_{i j}^{p}=\operatorname{softmax}(\sfrac{\bm{W}_{q}^{p} \bm{V}_{i}^{p}(\bm{W}_{k}^{p} \bm{V}_{j}^{p})^{T}}{\sqrt{d}})$
where $\bm{W}_{k}^{p}$ and $\bm{W}_{q}^{p}$, both $\in \mathbb{R}^{d\times d}$, are the weight matrices corresponding to the sending and receiving nodes, respectively. To allow the message-passing neural network to learn a diverse set of attention scores, we apply $H$ scaled dot-product self-attention heads in every message-passing step and concatenate their results. To this end, instead of using single weight matrices $\bm{W}_{q}^{p}, \bm{W}_{k}^{p}$ and $\bm{W}^{p}$, we use $\bm{W}_{q}^{p, h}, \bm{W}_{k}^{p, h}$ and $\bm{W}^{p, h}$ all $\in \mathbb{R}^{\sfrac{d}{H} \times d}$ for each attention head, resulting in:
\begin{equation}
\nonumber
\scriptsize
\bm{V}_{i}^{p+1}=\left[\sum_{j \in \mathcal{N}_{i}} \lambda_{i j}^{p, 1} \bm{W}^{p, 1} \bm{V}_{j}^{p}, \ldots, \sum_{j \in \mathcal{N}_{i}} \lambda_{i j}^{p, H} \bm{W}^{p, H} \bm{V}_{j}^{p}\right],
\end{equation} note that $\bm{V}_{i}^{p+1}$ still has the same dimension $\mathbb{R}^d$.

\subsection{Self-Supervised Pre-Training (\(\sampclr\))}
\label{ssec:self_sup}
The fact that we do not have access to the true class labels of the training data underscores the need to use a self-supervised pre-training scheme. As briefly discussed in \cref{sec:intro}, we build on the idea of employing contrastive prototypical transfer learning with some inspiration from \citep{Medina2020Self-SupervisedClassification, shirekar2022self, chen2020simple}.
Standard contrastive learning enforces embeddings of augmented images to be close to the embeddings of their source images in the representation space.
The \emph{key idea} of \texttt{SAMP-CLR} is not only to perform contrastive learning (the ``\texttt{CLR}'' component) on the source and augmented image embeddings, but also to ensure that images in the mini-batch belonging to potentially the same class have similar embeddings. This is where the ``\samp{}'' module comes to rescue, enabling the model to look beyond single instances and their augmentations. \samp{} allows the model to extract richer semantic information across multiple images present in a mini-batch.
Concretely speaking, we apply a contrastive loss on the \samp{} refined features (generated by $f_\psi$), and on the standard convolutional features (generated by $f_\theta$). Let us walk you through the process in more detail.

\begin{algorithm}[t]
    \caption{\texttt{SAMP-CLR}}\label{alg:samp-clr}
    \setstretch{1.35}
    \SetKwInOut{Input}{input}
    \SetKwInOut{Output}{output}
    \SetKwInput{Require}{Require}
	\SetKwInput{Return}{Return}
	\SetKw{Let}{let}
	\SetKwRepeat{Do}{do}{while}
	
	\SetAlgoLined
	\LinesNumbered
	\DontPrintSemicolon
	\SetNoFillComment
    \Require{$\mathcal{A}$, $f_\theta$, $f_\psi$, $\Omega$, $\alpha$, $\beta$, $\eta$, $d[.]$, $d^\prime[.]$}
    
    \While {not done}{
        Sample minibatch $\left\{\bm{x}_{i}\right\}_{i=1}^{L} \in \mathcal{D}_{\textup{tr}}$\;
        \text{Augment samples:} $\bar{\bm{x}}_{i, a}=\zeta_{a}(\bm{x}_{i})$; $\zeta_{a} \sim \mathcal{A}$.\;
        
        
		$\bm{Z}, \bm{\bar{Z}} \gets f_{\theta }\left(\left\{\bm{x}_{i}\right\}_{i=1}^{L}\right), f_{\theta }\left(\left\{\bar{\bm{x}}_{i,\ a}\right\}_{i=1,a=1}^{L,A}\right)$\;
                
        
        $\bm{V}= [\bm{Z}^\top, \bar{\bm{Z}}^\top]^\top$,  $\bm{e} = \{ d^\prime[\bm{V}_{i}, \bm{V}_{j}],  \forall i,j \in [B] \}$\;
        
        
        $\mathcal{G} \gets {\texttt{Graph}}(\bm{V}, \bm{e}, \gamma)$\;
        
         $\mathcal{G}^\prime \gets {\texttt{Graph}}(\bm{V}', \bm{e}', \gamma) = f_{\psi}(\mathcal{G})$

        
        $\bm{Z}^\prime, \bar{\bm{Z}}^\prime \gets \bm{V}^\prime_{1:L}, \bm{V}^\prime_{L+1:B}$ \;
        
        
        {\scalebox{1.0}{$\ell(i, a)=-\log \frac{\exp \left(-d\left[\bar{{\bm Z}}_{(a -1)L+i}, \bm{Z}_{i}\right]\right)}{\sum_{k=1}^{L} \exp \left(-d\left[\bar{{\bm Z}}_{(a -1)L+i},\bm{Z}_{k}\right]\right)}$}}\;
        
        
        $\mathscr{r}(i, a) = -\log \frac{\exp \left(-d\left[\bar{{\bm Z}}^{\prime}_{(a -1)L+i}, \bm{Z}^{\prime}_{i}\right]\right)}{\sum_{k=1}^{L} \exp \left(-d\left[\bar{{\bm Z}}^{\prime}_{(a -1)L+i},{\bm Z}^{\prime}_{k}\right]\right)}$ \;

        
        $\mathcal{L}_1=\sfrac{1}{L A} \sum_{i=1}^{L} \sum_{a=1}^{A} \ell(i, a)$\;
        $\mathcal{L}_2 = \sfrac{1}{L A} \sum_{i=1}^{L} \sum_{a=1}^{A} \mathscr{r}(i, a)$ \;
                
        
        $\mathcal{L} = \beta \mathcal{L}_1 + \mathcal{L}_2$\; 
        
        
        $\Omega \gets \Omega - \eta\nabla_{\Omega} \mathcal{L}$\;
    }
\end{algorithm}%

\Cref{alg:samp-clr} begins with batch generation: each mini-batch consists of $L$ random samples $\left\{\bm{x}_{i}\right\}_{i=1}^{L}$ from $\mathcal{D}_{\textup{tr}}$, where $\bm{x}_i$ is treated as a $1$-shot support sample for which we create $A$ randomly augmented versions $\tilde{\bm{x}}_{i, a}$ as query samples (lines $2$ to $3$), leading to a batch size of $B = (A + 1)L$.
Then the embeddings $\bm{Z} \in \mathbb{R}^{L\times d}$ and $\bar{\bm{Z}} \in \mathbb{R}^{L A\times d}$ are generated (line $4$) by passing the source images and augmented images through a feature extraction network $f_\theta$, respectively.
We then construct $\mathcal{G}=\texttt{Graph}(\bm{V}, \bm{e}, \gamma)$ with $\bm{V} = [\bm{Z}^\top, \bar{\bm{Z}}^\top]^\top$ of size $B \times d$ concatenating source and augmented image embeddings $\bm{Z}$ and $\bar{\bm{Z}}$ (line $5$-$6$), $\bm{e}$ is the vector of centered shift/scale-invariant cosine similarities $d'[.]$ (line $5$) \cite{van2012metric}, and $\gamma$ is defined earlier. 
The graph $\mathcal{G}$ is then passed through the \samp{} layer(s) $f_\psi$ resulting in a updated graph $\mathcal{G}'$ with refined node features $\bm{V}^\prime$ (line $7$). $\bm{V}'$ is spliced into the updated source and augmented image embeddings ($\bm{Z}^\prime$ and $\bar{\bm{Z}}^\prime$), respectively (lines $8$). In lines $9$ to $12$, we then apply contrastive losses $\mathcal{L}_1$ (between $\bm{Z}$ and $\bar{\bm{Z}}$) and $\mathcal{L}_2$ (between $\bm{Z}^\prime$ and $\bar{\bm{Z}}^\prime$).
Here, $\mathcal{L}_1$ encourages the feature extractor to cluster the embeddings of augmented query samples $\bar{\bm{Z}}$ around their prototypes (namely, source embeddings) $\bm{Z}$, which in turn provides a good initial set of embeddings for the \samp{} projector module to refine. $\mathcal{L}_2$ enforces the same constraints as $\mathcal{L}_1$ but for embeddings generated by the \samp{} layer.
Both loss terms use a Euclidean distance metric in the embedding space, denoted by $d[.]$. Finally, the overall loss is given by $\mathcal{L} = \beta\mathcal{L}_1 + \mathcal{L}_2$, which is optimized with mini-batch stochastic gradient descent w.r.t all the parameters in $\Omega = \{\theta, \psi\}$ where $\beta$ is a scaling factor, and $\eta$ the learning rate.

\subsection{Supervised Fine-tuning ($\opttune$)}
\label{ssec:sup_finetuning}
We propose a two-stage supervised fine-tuning consisting of (i) a transportation stage followed by (ii) a prototypical fine-tuning and classification stage.
The transportation stage involves using optimal transport (OT) \citep{cuturi2013sinkhorn, peyre2019computational}. As sketched in \cref{fig:classification}, OT helps projecting embeddings of the support set, $\bm{Z}^s=f_{\Omega }(\{\bm{x}^{s}_i\}_{i=1}^{NK}) \in \mathbb{R}^{NK \times d}$, so that they overlap better with the query set embeddings, $\bm{Z}^q = f_{\Omega }(\{\bm{x}^{q}_j\}_{j=1}^{NQ}) \in \mathbb{R}^{NQ \times d}$ upon transportation.
This increases the spread of $\bm{Z}^s$ in the query set's domain, which in turn creates more representative prototypes for each of the $N$ classes in $\mathcal{S}$. We show in \cref{sec:ablation-study} that this results in a significant boost in the downstream classification performance.

\textbf{OT based feature alignment.}
We provide a basic intuition for OT in the context of \ourmethod. Let $\bm{r} \in \mathbb{R}^{NK}$ and $\bm{c} \in \mathbb{R}^{NQ}$ be two probability simplexes defined over $\bm{Z}^{s}_i, \forall i \in [NK]$ and $\bm{Z}^{q}_j, \forall j \in [NQ]$, respectively. $\bm r$ denotes the distribution of the support embeddings, whereas $\bm c$ denotes the distribution of the query embeddings.
Consider $\bm{\Pi}(\bm{r}, \bm{c})$ to be a set of $NK \times NQ$ doubly stochastic matrices where all rows sum up to $\bm{r}$ and all columns sum up to $\bm{c}$ as:
\begin{equation*} \label{eqn:ot-def}
\footnotesize
\bm{\Pi}(\bm{r}, \bm{c})=\left\{\,\bm{\pi} \in \mathbb{R}_{+}^{NK \times NQ} \mid \bm{\pi} \mathbf{1}_{NQ}=\bm{r}, \bm{\pi}^{\top} \mathbf{1}_{NK}=\bm{c}\,\right\}.
\end{equation*}

\begin{algorithm}[t]
    \caption{$\opttune$}\label{alg:fine-tuning}
    \SetKwInOut{Input}{Input}
    \SetKwInOut{Output}{Output}
    \SetKwInput{Require}{Require}
	\SetKwInput{Return}{Return}
	\SetKw{Let}{let}
	\SetKwRepeat{Do}{do}{while}
	\SetAlgoLined
	\LinesNumbered
	\DontPrintSemicolon
	\SetNoFillComment
	
	\Require{$d[\cdot]$, $\bm{Z}^s$, $\bm{Z}^q$}

    $\bm{M}_{i,j}=d[\bm{Z}^{s}_i, \bm{Z}^{q}_j], \, \forall i \in [NK], j \in [NQ]$ \;
	$\bm{\pi}^{\star}$ $\gets$ Solving \cref{eqn:ot} using Sinkhorn-Knopp \citep{cuturi2013sinkhorn}\;
    $\hat{\bm{\pi}}^{\star}_{i, j} \leftarrow {\bm{\pi}^{\star}_{i, j}}\,/\,{\sum_{j} {\bm{\pi}}^{\star}_{i, j}}$, $\forall i \in [NK], j \in [NQ]$ \;
    Solve \cref{eqn:bcenter-map} \;
    \Return {$\hat{\bm{Z}}^s$}
\end{algorithm}

Intuitively, $\bm{\Pi}(\bm{r}, \bm{c})$ is a collection of all transport ``plans'', where a transport plan is defined as a potential strategy specifying how much of each support embedding is allocated to every query embedding and vice-versa. Our goal here is to find the most optimal transport plan, out of all possible transport plans $\bm{\Pi}(\bm{r}, \bm{c})$, that allocates $NK$ support embeddings to $NQ$ query embeddings with maximum overlap between their distributions. 

Given a cost matrix $\bm M$, the cost of mapping $\bm{Z}^s$ to $\bm{Z}^q$ using a transport plan $\bm{\pi}$ can be quantified as $\langle\bm{\pi}, \bm{M}\rangle_{F}$ and the OT problem can then be stated as,
\begin{equation}\label{eqn:ot}
\small
\bm{\pi}^{\star}=\underset{\bm{\pi} \in \bm{\Pi}\left(\bm{r}, \bm{c}\right)}{\operatorname{argmin}}\left\langle\bm{\pi}, \bm{M}\right\rangle_{F} - \varepsilon \mathbb{H}(\bm{\pi}),
\end{equation}
where $\bm{\pi}^\star$ denotes the most optimal transportation plan, $\langle \cdot, \cdot \rangle_{F}$ is the Frobenius dot product, and $\varepsilon$ is the weight on the entropic regularizer $\mathbb{H}$. The cost matrix $\bm{M}$ quantifies the overlap between the two distributions by measuring the distance between each support and query embedding pair: $\bm{M}_{i,j}=d[\bm{Z}^{s}_i, \bm{Z}^{q}_j]$.
The entropic regularization promotes \textquote{smoother} transportation plans \citep{cuturi2013sinkhorn}. \Cref{eqn:ot} is then solved using the time-efficient Sinkhorn-Knopp algorithm \citep{cuturi2013sinkhorn, sinkhorn1967concerning}. Notice that $\bm{\pi}^\star$ is also referred to as \textit{Wasserstein metric} \citep{peyre2019computational, cuturi2013sinkhorn}. To adapt $\bm{Z}^s$ to $\bm{Z}^q$ with cost matrix $\bm{M}$, we compute $\hat{\bm{Z}}^s$  as the \textit{projected mapping} of ${\bm{Z}^s}$, given by:
\begin{equation}
\begin{aligned}\label{eqn:bcenter-map}
\hat{\bm{Z}}^s &= \hat{\bm{\pi}}^{\star} \bm{Z}^q ,\\
\hat{\bm{\pi}}^{\star}_{i, j} = \frac{\bm{\pi}^{\star}_{i, j}}{\sum_{j} {\bm{\pi}}^{\star}_{i, j}} &\,, \forall i \in [NK], j \in [NQ], 
\end{aligned}
\end{equation}
\belowdisplayshortskip=0pt
\noindent where $\widehat{\bm{\pi}}^{\star}$ is the normalized transport.
The \textit{projected support} embeddings $\hat{\bm{Z}}^s$ are an estimation of $\bm{Z}^s$ in the region occupied by the query embeddings $\bm{Z}^q$.
Specifically, it is a barycentric mapping of the support features $\bm{Z}^s$. \Cref{alg:fine-tuning} shows this process in a succinct manner.

{\bf Prototypical classification}. The projected support embeddings, $\hat{\bm{Z}}^s$, are used for prototype creation and classification of the query points. To this end, following \citep{Medina2020Self-SupervisedClassification,Triantafillou2020Meta-Dataset:Examples} we concatenate $f_\Omega$ with a single layer nearest mean classifier $f_\phi$ (resulting in an architecture similar to ProtoNet \citep{Snell2017PrototypicalLearning}) and only fine-tune this last layer. In this stage, for each class $k \in \mathcal{C}$ in the support set, we compute the class prototype $\bm{c}_k$ for class $k$ using the projected support embeddings $\hat{\bm{Z}}^{s,k}$ belonging to class $k$:
\begin{equation*}
\abovedisplayskip=0pt
\bm{c}_{k} = \frac{1}{\left\vert\hat{\bm{Z}}^{s,k} \right\vert} \sum_{\hat{\bm{z}} \in \hat{\bm{Z}}^{s,k}}\hat{\bm{z}} \text{, for } k \in \mathcal{C}.
\end{equation*}
Following \citep{Triantafillou2020Meta-Dataset:Examples,Medina2020Self-SupervisedClassification}, we initialize the classification layer $f_\phi$ with weights set to $\bm{W}_k = 2\bm{c}_k$ and biases set to $b_{k}=-\|\bm{c}_{k}\|^{2}$. To finetune this layer, we sample a subset of supports from $\mathcal{S}$ and train $f_\phi$ with a standard cross-entropy loss; more details are given in \cref{sec:experimental-setup}.

\section{Experimental Setup}
\label{sec:experimental-setup}
\textbf{Datasets.}\label{sssec:datasets}
To benchmark the performance of our method \ourmethod{}, we conduct ``in-domain'' experimentation on two most commonly adopted few-shot learning datasets: \miniImagenet{} \citep{Vinyals2016MatchingLearning} and \tieredImagenet{} \citep{ren2018meta}. \textit{Mini}ImageNet contains $100$ classes with $600$ samples in each class. This equals a total of $60,000$ images that we resize to $84 \times 84$ pixels. Out of the $100$ classes, we use $64$ classes for training, $16$ for validation, and $20$ for testing.
\textit{Tiered}ImageNet is a larger subset of ILSVRC-12 \citep{deng2009imagenet} with $608$ classes with a total of $779,165$ images of size $84 \times 84$. We use $351$ for training, $97$ for validation, and $8$ for testing, out of the $608$ classes. The augmentation strategy follows the one proposed in \citep{bachman2019learning}. We also compare our method on a recent more challenging ``cross-domain'' few-shot learning (CDFSL) benchmark \citep{guo2019new}, which consists of several datasets. This benchmark has four datasets with increasing similarities to \miniImagenet. In that order, we have grayscale chest X-ray images from ChestX \citep{wang2017chestx}, dermatological skin lesion images from ISIC2018 \cite{isic2018dataset}, aerial satellite images from EuroSAT \citep{helber2017eurosat}, and crop disease images from CropDiseases \citep{mohanty2016using}. 
We also used the Caltech-UCSD Birds (CUB) dataset \citep{WahCUB_200_2011} for further analysis of cross-domain performance. The CUB dataset is made up of $11,788$ images from $200$ unique species of birds. We use $100$ classes for training, $50$ for both validation and testing.

\textbf{Training strategy.}
In \cref{fig:pre-training}, as feature extractor, we use the standard \cfs{} model following \citep{Vinyals2016MatchingLearning, Medina2020Self-SupervisedClassification, Khodadadeh2018UnsupervisedClassification}. It is followed by a single \samp{} layer with $4$ attention heads. Note that we also use a slightly modified version of the \cfs{} network which we call \cfb{}, where we increase the number of filters from $(64, 64, 64, 64)$ to $(96, 128, 256, 512)$ \citep{gidaris2019boosting} and average pool the final feature map returning a smaller embedding dimension $d=512$ instead of $d=1600$.
The networks are pre-trained using $\sampclr$ on the respective training splits of the datasets, with an initial learning rate of $\eta = 0.0005$, annealed by a cosine scheduler via the Adam optimizer \citep{kingma2014adam} and $L=128$.
Experiments involving CDFSL benchmark follow \citep{guo2019new, Medina2020Self-SupervisedClassification, shirekar2022self}, where we pre-train a ResNet-$10$ encoder using $\sampclr$ on \miniImagenet{} images of size $224 \times 224$ for $400$ epochs with the Adam optimizer and a constant learning rate of $\eta = 0.0001$. Similar to the \cfs{} encoder, the ResNet-$10$ uses the same \samp{} configuration.
During validation and testing, as defined in \cref{ssec:sup_finetuning}, we initialize and fine-tune $f_\phi$ for $15$ iterations where we sample a subset of examples from $\mathcal{S}$ in each iteration. For validation, we create $15$ ($N$-way, $K$-shot) tasks using the validation split of the respective dataset.

\textbf{Evaluation scenarios and baseline.}
Our testing scheme uses $600$ test episodes, each with $15$ query shots per class, on which the pre-trained encoder ($\sampclr$) is fine-tuned using \(\opttune\) and tested. All our results indicate $95\%$ confidence intervals over $3$ runs, each with $600$ test episodes. Therefore, the standard deviation values are calculated according to the $3$ runs to provide more concrete measures for comparison. For our in-domain benchmarks, we test on ($5$-way, $1$-shot) and ($5$-way, $5$-shot) classification tasks, while our cross-domain testing is performed using ($5$-way, $5$-shot) and ($5$-way, $20$-shot) classification tasks following \citep{guo2019new}.
We compare our performance with a suite of recent unsupervised few-shot baselines such as U-MlSo \citep{zhang2022miso}, C$^3$LR \citep{shirekar2022self}, Meta-GMVAE \citep{lee2021meta}, and Revisiting UML \citep{Ye2022} to name a few. Furthermore, we also compare with a set of supervised approaches (such as MetaQDA \citep{Zhang_2021_ICCV} and TransductiveCNAPS \citep{bateni2022enhancing}), the best of which are expected to outperform ours and other unsupervised methods.

\begin{table}[t!]
	\footnotesize
	\caption{Accuracy ($\% \pm$ std.) for ($N$-way, $K$-shot) classification tasks. Style: \textbf{best} and \underline{second best}.}
	\vspace{-0.2cm}
		\centering
		{\tabcolsep=0pt\def\arraystretch{1.1}
			\begin{tabularx}{\linewidth}{l  *3{@{}>{\centering\arraybackslash}X}}
				\toprule
				& &\multicolumn{2}{c}{\bf \miniImagenet}
				\tabularnewline \cmidrule(lr){3-4}
				{\bf Method$(N,K)$}         & \textbf{Backbone}                                  & {(5,1)}                             & {(5,5)}                             \\ 
				        
				\midrule
				        
				CACTUs-MAML \citep{Hsu2018UnsupervisedMeta-Learning}           &    Conv4             &39.90 \scriptsize{$\pm$ 0.74}             & 53.97 \scriptsize{$\pm$ 0.70}             \\
				CACTUs-Proto \cite{Hsu2018UnsupervisedMeta-Learning}           &    Conv4             &39.18 \scriptsize{$\pm$ 0.71}             & 53.36 \scriptsize{$\pm$ 0.70}             \\
				UMTRA \citep{Khodadadeh2018UnsupervisedClassification}         &    Conv4             &39.93                               & 50.73                               \\
				AAL-ProtoNet \citep{Antoniou2019AssumeAugmentation}            &    Conv4             &37.67 \scriptsize{$\pm$ 0.39}             & 40.29 \scriptsize{$\pm$ 0.68}             \\
				AAL-MAML++ \citep{Antoniou2019AssumeAugmentation}              &    Conv4             &34.57 \scriptsize{$\pm$ 0.74}             & 49.18\scriptsize{$\pm$ 0.47}              \\
				UFLST \citep{Ji2019UnsupervisedTraining}                       &    Conv4             &33.77 \scriptsize{$\pm$ 0.70}             & 45.03 \scriptsize{$\pm$ 0.73}             \\
				ULDA-ProtoNet \citep{Qin2020DiversityAugmentation}             &    Conv4             &40.63 \scriptsize{$\pm$ 0.61}             & 55.41 \scriptsize{$\pm$ 0.57}             \\
				ULDA-MetaNet \citep{Qin2020DiversityAugmentation}              &    Conv4             &40.71 \scriptsize{$\pm$ 0.62}             & 54.49 \scriptsize{$\pm$ 0.58}             \\
				U-SoSN+ArL \citep{Zhang2020RethinkingLearning}                 &    Conv4             &41.13 \scriptsize{$\pm$ 0.84}             & 55.39 \scriptsize{$\pm$ 0.79}             \\
				U-MlSo \citep{zhang2022miso}                                   &    Conv4             &41.09                               & 55.38                               \\
				ProtoTransfer \citep{Medina2020Self-SupervisedClassification}  &    Conv4             &45.67 \scriptsize{$\pm$ 0.79}             & 62.99 \scriptsize{$\pm$ 0.75}             \\
				CUMCA \citep{xu2021unsupervised}                               &    Conv4             &41.12                               & 54.55                               \\
				Meta-GMVAE \citep{lee2021meta}                                 &    Conv4             &42.82                               & 55.73                               \\
				Revisiting UML \citep{Ye2022}                                  &    Conv4             &48.12 \scriptsize{$\pm$ 0.19}             & \underline{65.33 \scriptsize{$\pm$ 0.17}} \\
				CSSL-FSL\_Mini64 \citep{li2020few}                             &    Conv4             &\underline{48.53 \scriptsize{$\pm$ 1.26}} & 63.13 \scriptsize{$\pm$ 0.87}             \\
				C$^3$LR \citep{shirekar2022self}                               &    Conv4             &47.92 \scriptsize{$\pm$ 1.2}              & 64.81 \scriptsize{$\pm$ 1.15}             \\
				\rowcolor{teal!10} $\ourmethod$ (\textbf{ours})                                        &    Conv4             &55.75 \scriptsize{$\pm$ 0.77}    & {68.33} \scriptsize{$\pm$ 0.66}    \\
				\rowcolor{teal!20} $\ourmethod^\star$ (\textbf{ours})                     &    Conv4b         &\textbf{61.02} \scriptsize{$\pm$ 1.0}              & \textbf{72.52} \scriptsize{$\pm$ 0.68}             \\
				\cdashlinelr{1-4}
				{\em Supervised Methods} \\
				{MAML \citep{Finn2017Model-agnosticNetworks}}                 & Conv4        &46.81 \scriptsize{$\pm$ 0.77}             & 62.13\scriptsize{$\pm$ 0.72}              \\
				{ProtoNet \citep{Snell2017PrototypicalLearning}}               & Conv4        &46.44\scriptsize{$\pm$ 0.78}              & 66.33\scriptsize{$\pm$ 0.68}              \\
				{MMC \citep{ren2018meta} }                                    & Conv4        &50.41 \scriptsize{$\pm$ 0.31}             & 64.39 \scriptsize{$\pm$ 0.24}             \\
				{FEAT \citep{ye2020few}}                                       & Conv4        &55.15                               & 71.61                               \\
				SimpleShot \citep{wang2019simpleshot}                          & Conv4        &49.69 \scriptsize{$\pm$ 0.19}             & 66.92 \scriptsize{$\pm$ 0.17}             \\
				Simple CNAPS \citep{bateni2022enhancing}                       & ResNet-18        &53.2 \scriptsize{$\pm$ 0.9}               & 70.8 \scriptsize{ $\pm$ 0.7}              \\
				Transductive CNAPS \citep{bateni2022enhancing}                 & ResNet-18 & 55.6 \scriptsize{$\pm$ 0.9} & 73.1 \scriptsize{$\pm$ 0.7} \\
				MetaQDA \citep{Zhang_2021_ICCV}                                & Conv4        &56.41 \scriptsize{$\pm$ 0.80}             & 72.64 \scriptsize{$\pm$ 0.62}             \\
				{Pre+Linear \citep{Medina2020Self-SupervisedClassification}}  & Conv4        &43.87 \scriptsize{$\pm$ 0.69}             & 63.01 \scriptsize{$\pm$ 0.71}             \\
				\bottomrule
			\end{tabularx}}
	\vspace{-0.2cm}
	\label{tab:ResultsMini}
\end{table}

\begin{table}[t!]
	\footnotesize
    \caption{Accuracy ($\% \pm$ std.) for ($N$-way, $K$-shot) classification tasks. Style: \textbf{best} and \underline{second best}.}
    \vspace{-0.2cm}
    \centering
	{\tabcolsep=0pt\def\arraystretch{1.1}
		\begin{tabularx}{\linewidth}{l  *3{@{}>{\centering\arraybackslash}X}}
        \toprule
        & & \multicolumn{2}{c}{\bf \tieredImagenet}
        \tabularnewline \cmidrule(lr){3-4}
        {\bf Method$(N,K)$} & \textbf{Backbone} & {(5,1)} & {(5,5)} \\ 
        
        \midrule
        
    	C$^3$LR \citep{shirekar2022self} & Conv4 & 42.37 \scriptsize{$\pm$ 0.77}  & \underline{61.77} \scriptsize{$\pm$ 0.25} \\
    	
    	ULDA-ProtoNet \citep{Qin2020DiversityAugmentation}& Conv4 & 41.60 \scriptsize{$\pm$0.64} & 56.28 \scriptsize{$\pm$ 0.62} \\
    	ULDA-MetaOptNet \citep{Qin2020DiversityAugmentation} & Conv4 & 41.77 \scriptsize{$\pm$ 0.65} & 56.78 \scriptsize{$\pm$ 0.63} \\
    	U-SoSN+ArL \citep{Zhang2020RethinkingLearning} & Conv4 & \underline{43.68} \scriptsize{ $\pm$ 0.91} & 58.56 \scriptsize{ $\pm$ 0.74} \\
    	U-MlSo \citep{zhang2022miso} & Conv4 & 43.01 \scriptsize{$\pm$ 0.91} & 57.53 \scriptsize{$\pm$ 0.74} \\
    	\rowcolor{teal!10} $\ourmethod$ (\textbf{ours}) & Conv4 & 45.25 \scriptsize{$\pm$ 0.89} & 59.75 \scriptsize{$\pm$ 0.66} \\
    	
    	\rowcolor{teal!20} $\ourmethod^\star$ (\textbf{ours})& Conv4b & \textbf{49.10} \scriptsize{$\pm$ 0.94} & \textbf{65.19}	\scriptsize{$\pm$ 0.82} \\
    	
        \bottomrule
        \end{tabularx}}
        \vspace{-0.4cm}
    \label{tab:ResultsTiered}
\end{table}
\begin{table*}[t!]
	\footnotesize
	\centering
	\caption{Accuracy ($\% \pm$ std.) of ($N$-way, $K$-shot) classification on the CDFSL benchmark. Style: \textbf{best} and \underline{second best}.}
	\vspace{-0.2cm}
	\begin{adjustbox}{width=\textwidth}
		{\tabcolsep=0pt\def\arraystretch{1.0}
			\begin{tabularx}{\linewidth}{@{}>{\arraybackslash}l| *2{@{}>{\centering\arraybackslash}X} | *2{>{\centering\arraybackslash}X} | *2{>{\centering\arraybackslash}X} | *2{>{\centering\arraybackslash}X}}
				\toprule
				{\bf Method$(N,K)$}                                             & {(5,5)}                                   & {(5,20)}                                  & {(5,5)}                                   & {(5,20)}                                  & {(5,5)}                                   & {(5,20)}                                  & {(5,5)}                                   & {(5,20)}                                  \\ 
				\midrule
								            
				& \multicolumn{2}{c}{\cellcolor[HTML]{BEBDFF}\bf ChestX} & \multicolumn{2}{|c}{\cellcolor[HTML]{FAE0C1}\bf{ISIC}} & \multicolumn{2}{|c}{\cellcolor[HTML]{E7F2F8}\bf EuroSAT}  & \multicolumn{2}{|c}{\cellcolor[HTML]{B4F8C8}\bf CropDiseases} \\
								            
				\midrule
				UMTRA-ProtoNet \citep{Medina2020Self-SupervisedClassification}  & 24.94 \scriptsize{$\pm$ 0.43}             & 28.04 \scriptsize{$\pm$ 0.44}             & 39.21 \scriptsize{$\pm$ 0.53}             & 44.62 \scriptsize{$\pm$ 0.49}             & 74.91 \scriptsize{$\pm$ 0.72}             & 80.42 \scriptsize{$\pm$ 0.66}             & 79.81 \scriptsize{$\pm$ 0.65}             & 86.84 \scriptsize{$\pm$ 0.50}             \\
				UMTRA-ProtoTune \citep{Medina2020Self-SupervisedClassification} & 25.00 \scriptsize{$\pm$ 0.43}             & 30.41 \scriptsize{$\pm$ 0.44}             & 38.47 \scriptsize{$\pm$ 0.55}             & 51.60 \scriptsize{$\pm$ 0.54}             & 68.11 \scriptsize{$\pm$ 0.70}             & 81.56 \scriptsize{$\pm$ 0.54}             & 82.67 \scriptsize{$\pm$ 0.60}             & 92.04 \scriptsize{$\pm$ 0.43}             \\
				ProtoTransfer \citep{Medina2020Self-SupervisedClassification}   & \underline{26.71} \scriptsize{$\pm$ 0.46}    & 33.82 \scriptsize{$\pm$ 0.48}    & 45.19 \scriptsize{$\pm$ 0.56} & 59.07 \scriptsize{$\pm$ 0.55} & 75.62 \scriptsize{$\pm$ 0.67} & 86.80 \scriptsize{$\pm$ 0.42} & 86.53 \scriptsize{$\pm$ 0.56} & 95.06 \scriptsize{$\pm$ 0.32} \\
				C$^3$LR \citep{shirekar2022self} & 26.00 \scriptsize{$\pm$ 0.41} & 33.39 \scriptsize{$\pm$ 0.47} & 45.93 \scriptsize{$\pm$ 0.54}    & 59.95 \scriptsize{$\pm$ 0.53}    & 80.32 \scriptsize{$\pm$ 0.65}    & 88.09 \scriptsize{$\pm$ 0.45}    & 87.90 \scriptsize{$\pm$ 0.55}    & 95.38 \scriptsize{$\pm$ 0.31}    \\
				\rowcolor{teal!20} $\ourmethod$ (\textbf{ours})                       & {26.27} \scriptsize{$\pm$ 0.44} & \textbf{34.15} \scriptsize{$\pm$ 0.50} & \underline{47.60} \scriptsize{$\pm$ 0.59}    & \textbf{61.28} \scriptsize{$\pm$ 0.56}    & \textbf{85.55} \scriptsize{$\pm$ 0.60}    & \underline{88.52} \scriptsize{$\pm$ 0.50}    & \textbf{91.74} \scriptsize{$\pm$ 0.55}    & \textbf{96.36} \scriptsize{$\pm$ 0.28}    \\
								
				\cdashlinelr{1-9}
				
				ConFeSS \citep{das2022confess} (dedicated) & \textbf{27.09} & \underline{33.57} & \textbf{48.85} & \underline{60.10} & \underline{84.65} & \textbf{90.40} & \underline{88.88} & \underline{95.34} \\
				
				ATA \citep{Wang2021} (dedicated) & 24.43 \scriptsize{$\pm$ 0.2} & - & 45.83 \scriptsize{ $\pm$ 0.3} & - & 83.75 \scriptsize{$\pm$ 0.4} & - & 90.59 \scriptsize{$\pm$ 0.3} & - \\
				
				
				ProtoNet \citep{guo2019new} (sup.)                              & 24.05 \scriptsize{$\pm$ 1.01}             & 28.21 \scriptsize{$\pm$ 1.15}             & 39.57 \scriptsize{$\pm$ 0.57}             & 49.50 \scriptsize{$\pm$ 0.55}             & 73.29 \scriptsize{$\pm$ 0.71}             & 82.27 \scriptsize{$\pm$ 0.57}             & 79.72 \scriptsize{$\pm$ 0.67}             & 88.15 \scriptsize{$\pm$ 0.51}             \\
				Pre+Mean-Cent. \citep{guo2019new} (sup.)                        & 26.31 \scriptsize{$\pm$ 0.42}             & 30.41 \scriptsize{$\pm$ 0.46}             & 47.16 \scriptsize{$\pm$ 0.54}             & 56.40 \scriptsize{$\pm$ 0.53}             & 82.21 \scriptsize{$\pm$ 0.49}             & 87.62 \scriptsize{$\pm$ 0.34}             & 87.61 \scriptsize{$\pm$ 0.47}             & 93.87 \scriptsize{$\pm$ 0.68}             \\
				Pre+Linear \citep{guo2019new} (sup.)                            & 25.97 \scriptsize{$\pm$ 0.41}             & 31.32 \scriptsize{$\pm$ 0.45}             & 48.11 \scriptsize{$\pm$ 0.64}             & 59.31 \scriptsize{$\pm$ 0.48}             & 79.08 \scriptsize{$\pm$ 0.61}             & 87.64 \scriptsize{$\pm$ 0.47}             & 89.25 \scriptsize{$\pm$ 0.51}             & 95.51 \scriptsize{$\pm$ 0.31}             \\
				

				\bottomrule
			\end{tabularx}}
	\end{adjustbox}
	\vspace{-0.4cm}
	\label{tab:cdfsl_full}
		
\end{table*}
\begin{table}[t]
        \footnotesize
        \centering
        \caption{Ablation study of various parameters on accuracy.}
        \vspace{-0.2cm}
        {\tabcolsep=1pt\def\arraystretch{1}
        \begin{tabularx}{\columnwidth}{l *6{@{\hspace{-5pt}}>{\centering\arraybackslash}X}}
        \toprule
            \textbf{Backbone}& $p$ & $H$ & $L$  & $\beta$ & \textbf{OT} & \textbf{Accuracy} \\
            \midrule
            Conv4b    & 1 & 4 & 64 &  1.0 & \cmark  &  71.42 \tiny{$\pm$ 0.73} \\
            Conv4b    & 1 & 4 & 64 &  0.7 & \cmark  &  71.41 \tiny{$\pm$ 0.71} \\
            Conv4b    & 1 & 8 & 64 &  1.0 & \cmark  &  71.27 \tiny{$\pm$ 0.75} \\
            Conv4b    & 1 & 8 & 64 &  0.7 & \cmark  &  69.87 \tiny{$\pm$ 0.72} \\
            Conv4b    & 2 & 1 & 64 &  0.7 & \cmark  &  68.99 \tiny{$\pm$ 0.71} \\ 
            Conv4b    & 2 & 4 & 64 &  0.7 & \cmark  &  67.01 \tiny{$\pm$ 0.69} \\
            
            Conv4     & 1 & 4 & 64 &  0.7 & \cmark  &  69.61 \tiny{$\pm$ 0.71} \\
            Conv4     & 1 & 4 & 64 &  1.0 & \cmark  &  67.60 \tiny{$\pm$ 0.62} \\ 
            Conv4     & 1 & 8 & 64 &  1.0 & \cmark  &  63.59 \tiny{$\pm$ 0.68} \\ 
            
            Conv4b    & 1 & 4 & 128 & 0.7 & \cmark  &  72.52 \tiny{$\pm$ 0.72} \\ 
            Conv4     & 1 & 4 & 128  & 0.7 & \cmark  &  68.33 \tiny{$\pm$ 0.71} \\ 
            Conv4     & 1 & 4 & 128 &  0.0  & \cmark  &  52.81 \tiny{$\pm$ 0.66} \\
            Conv4b    &	1 &	4 &	128 & 0.0 & \cmark  &  72.44 \tiny{$\pm$ 0.69} \\
            \cdashlinelr{1-7}
            Conv4b    & 1 &	4 &	64  &  0.7 & \xmark  &  64.29 \tiny{$\pm$ 0.63} \\
            Conv4b    &	1 &	4 &	128 & 0.7 & \xmark  &  63.47 \tiny{$\pm$ 0.64} \\
            Conv4     & 1 & 4 & 64 &  0.7  & \xmark  &  66.73 \tiny{$\pm$ 0.65} \\
            
        \bottomrule
        \end{tabularx}}\label{tab:conv4-ablation} 
        \vspace{-0.2cm}
\end{table}

\begin{table}[t]
    \footnotesize
    \caption{Accuracy ($\% \pm$ std.) for ($N$-way, $K$-shot) classification on \miniImagenet{} with pre-training on CUB.}
    \vspace{-0.2cm}
    {\tabcolsep=0pt\def\arraystretch{1}
    \begin{tabularx}{\linewidth}{l *3{@{}>{\centering\arraybackslash}X}}
        \toprule
        {\bf Training} & {\bf Testing} & {(5,1)} & {(5,5)} \\
        \midrule
        ProtoTransfer \small{\citep{Medina2020Self-SupervisedClassification}} & ProtoTune  & 35.37 \scriptsize{$\pm$ 0.63} & 52.38 \scriptsize{$\pm$ 0.66} \\
        C$^3$LR \citep{shirekar2022self} & ProtoTune & \underline{39.61} \scriptsize{$\pm$ 1.11} & \underline{55.53} \scriptsize{$\pm$ 1.42} \\
        \rowcolor{teal!20} $\ourmethod$ (\textbf{ours}) & OpT-Tune & \textbf{49.32} \scriptsize{$\pm$ 0.75} & \textbf{56.10} \scriptsize{$\pm$ 0.60} \\
        \bottomrule
        \end{tabularx}}
        \vspace{-0.4cm}
    \label{tab:results_cub2mini}
\end{table}

\section{Performance Evaluation}
\label{sec:perf-eval}
\textbf{In-domain experiments.}
\Cref{tab:ResultsMini} summarizes our performance evaluation results on the \miniImagenet{} dataset for ($N$-way, $K$-shot) scenarios with $N = 5$ and $K = 1, 5$. The top section compares the performance of the proposed approach (\ourmethod{}) with the most recent unsupervised competitors. 
We outperform our closest competitors by approximately $7\%+$ and $2\%+$ in the ($5$-way, $1$-shot) and ($5$-way, $5$-shot) settings, respectively. 
More interestingly, our method matches or outperforms some of the supervised baselines (bottom section of the table), especially SimpleCNAPS which uses a much more powerful ResNet-$18$ backbone. Obviously, the state-of-the-art supervised few-shot learning approaches have the advantage of having access to the true labels.
When it comes to \tieredImagenet{}, our approach shows considerable gains over recent competitors such as C$^3$LR \citep{shirekar2022self} with a $3\%+$ improvement in the ($5$-way, $1$-shot) setting and a $5\%+$ improvement in the ($5$-way,$5$ shot) setting. As such, \ourmethod{} sets a new state-of-the-art for both \tieredImagenet{} and \miniImagenet{} datasets.

\textbf{Cross-domain experiments.}
 We focus on the recent CDFSL benchmark \citep{guo2019new} to investigate the performance of $\ourmethod$ in cross-domain scenarios. This outcome is summarize in \Cref{tab:cdfsl_full}. 
Here, we pre-train on \miniImagenet\ and fine-tune on ChestX \citep{wang2017chestx}, ISIC2018 \citep{isic2018dataset}, EuroSAT \citep{helber2017eurosat}, and CropDiseases \citep{mohanty2016using}.
We compare the performance against C$^3$LR\citep{shirekar2022self}, ProtoTransfer \citep{Medina2020Self-SupervisedClassification} along with its two variants using UMTRA \citep{Khodadadeh2018UnsupervisedClassification} (also proposed in \citep{Medina2020Self-SupervisedClassification}), as well as ConFeSS \citep{das2022confess} and ATA \citep{Wang2021} - two of the latest methods \emph{dedicated} to solving the cross-domain few-shot learning problem.
Note that we also compare with a couple of related supervised approaches from \citep{guo2019new}, as a reference. Our method consistently keeps up with ConFeSS \citep{das2022confess}, but scores higher in $5$ and $20$ shot CropDiseases tasks by $2\%+$ and about $1\%$, respectively. Except for EuroSAT, our method is consistently competitive ($\sim1\%$ difference) to the performance of ConFeSS in ChestX and ISIC. In ISIC, which is the second least similar dataset to \miniImagenet, our method is better by $1\%+$ in the ($5$-way, $20$-shot) setting. 
Note that \ourmethod\ outperforms another recent dedicated method ATA \citep{Wang2021} in all but one CDFSL benchmark settings, with the exception being the EuroSAT ($5$-way, $5$-shot) setting.

\vspace{-0.2cm}
\section{Ablation Study and Robustness Analysis}
\vspace{-0.1cm}
\label{sec:ablation-study}
\Cref{tab:conv4-ablation} investigates the performance of the proposed method against various choices of important hyperparameters. We use the ($5$-way, $5$-shot) \miniImagenet{} benchmark to analyze the robustness of our method and demonstrate the importance of our design choices.
\begin{figure}[t!]
\begin{minipage}[t]{0.49\linewidth}
    \captionsetup{font=small}
    \includegraphics[width=\linewidth, trim=3.2cm 3cm 2.6cm 2.6cm, clip]{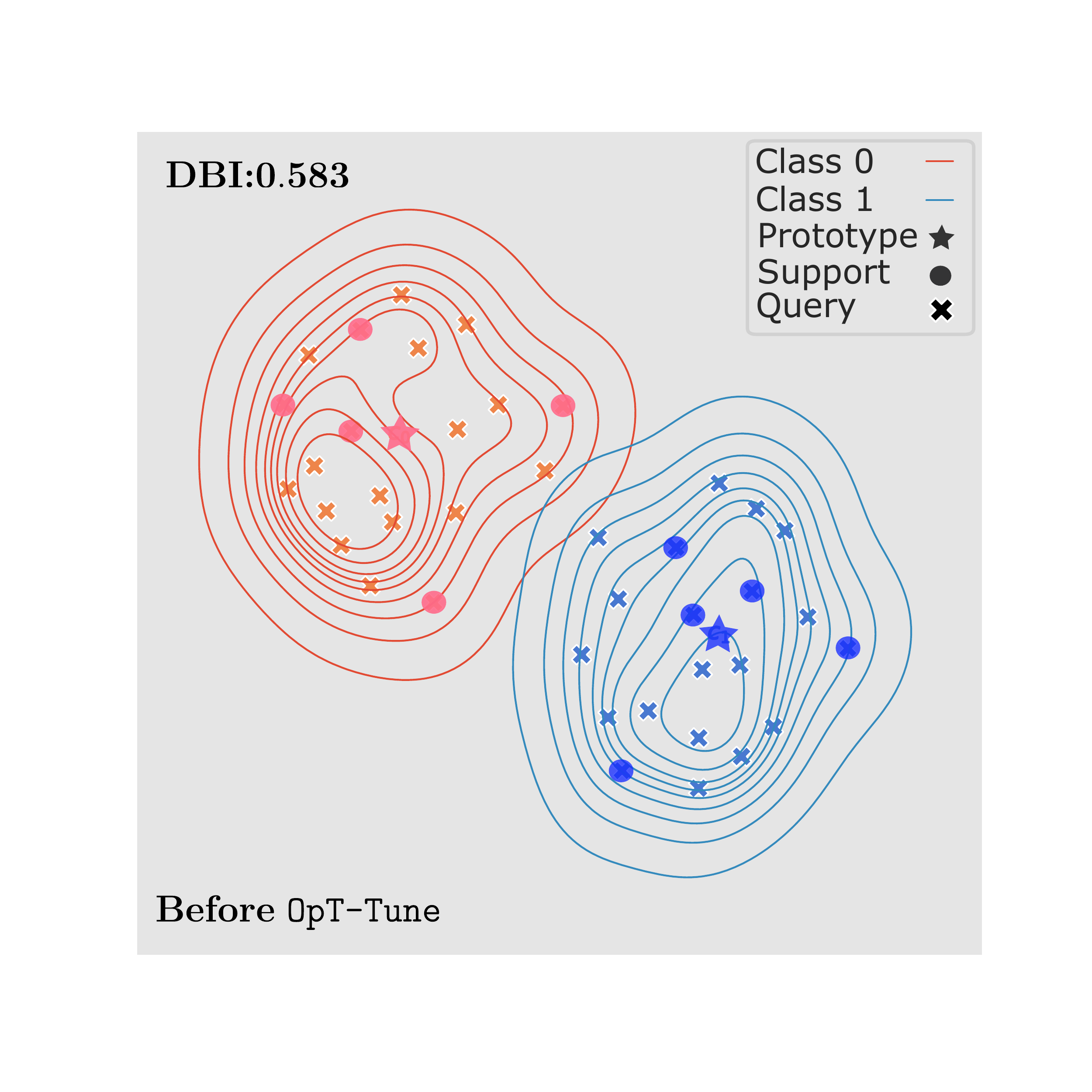}
    \label{fig:raw_emb}
    \vspace{-0.5cm}
\end{minipage}%
    \hfill%
\begin{minipage}[t]{0.49\linewidth}
    \captionsetup{font=small}
    \includegraphics[width=\linewidth, trim=3.2cm 3cm 2.6cm 2.6cm, clip]{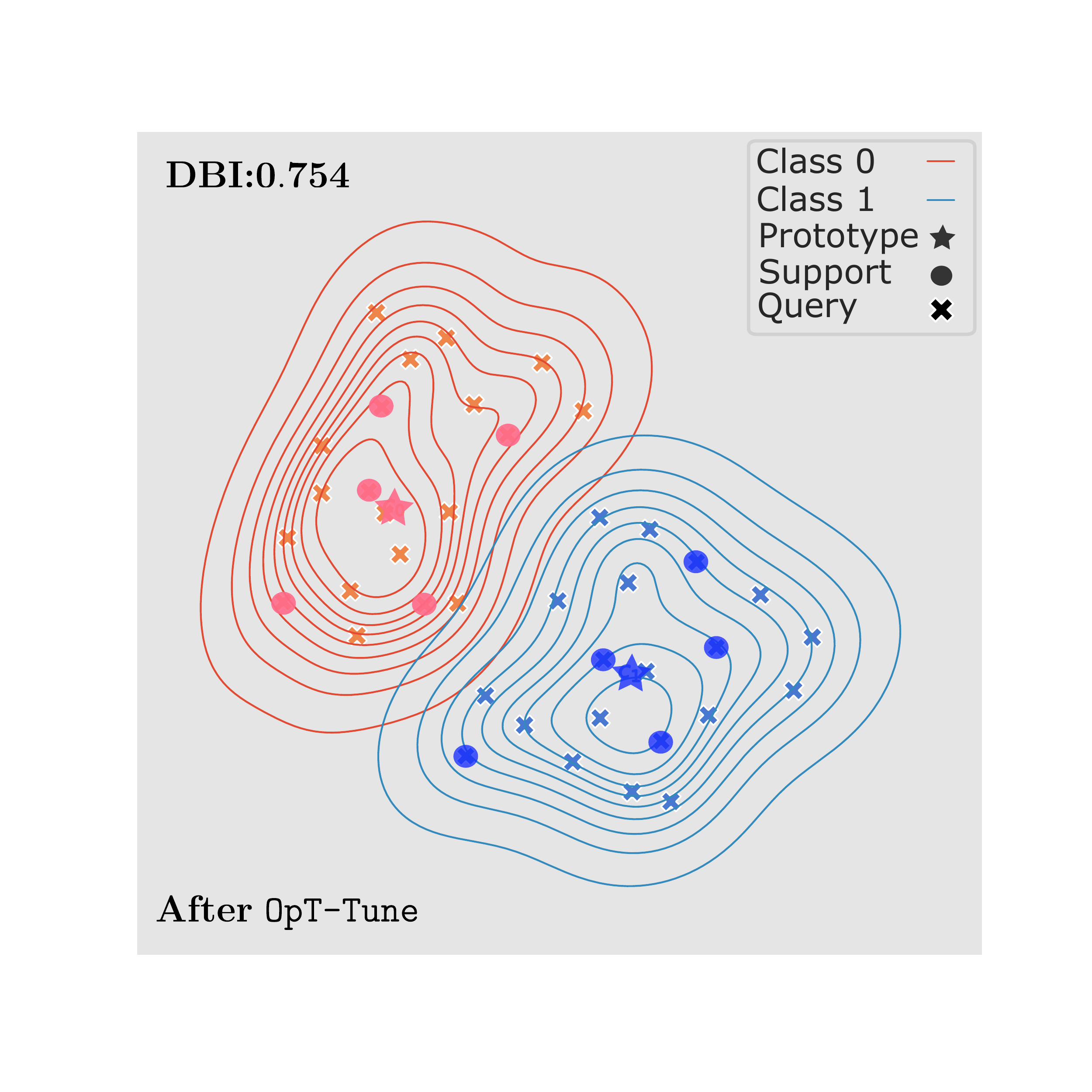}
    \label{fig:after-ot}
    \vspace{-0.5cm}
\end{minipage} 
    \caption{Before (left) and after applying OT (right). Prototypes (\ding[0.8]{72}), supports (\ding[0.8]{108}) and queries (\ding[0.8]{54}). OT helps better align the distribution of support and query samples.}
    \label{fig:ot-application}
    \vspace{-0.4cm}
\end{figure}

\textbf{\opttune\ is crucial.} To illustrate the effect of using $\opttune$ on the classification performance, we perform experiments with $\opttune$ disabled. For a fair comparison, we use the same pre-trained models in the test runs with $\opttune$ enabled or disabled.
The best performing model (a \cfb{}) uses $1$ \samp{} layer with $4$ attention heads and a batch size of $128$, resulting in accuracy of $72.52\%$ with $\opttune$ enabled. The same model, with $\opttune$ disabled, loses $9\%$ accuracy.
Even with $\opttune$ disabled, our method remains competitive with some of the latest methods in \cref{tab:ResultsMini}. This observation suggests that the process described in \cref{ssec:sup_finetuning} is an efficient technique to incorporate task awareness and improve the quality of prototypes. This is further corroborated in \cref{fig:ot-application} where a task with $N=2$ is used to showcase the effect of \opttune{}. We observe that the support embeddings are more evenly spread out over the distribution of the query embeddings. This is also backed by the DBI score \citep{davie1979} which increases from $0.583$ to $0.754$ after \opttune{} is applied.

\textbf{SAMP layers and attention heads.}
In \cref{tab:conv4-ablation}, we also investigate the robustness of our method when the number of \samp{} layers ($p$) and attention heads ($H$) vary. The best performance is achieved with a single \samp{} layer with four attention heads. 
Increasing $p$ leads to a significant decrease in performance; however, increasing $H$ leads to a small performance degradation.
Notably, the observations here are consistent with those reported in \citep{velic018graph, seidenschwarz2021learning}.

\textbf{Embedding dimension.} We measure the performance of the model in relation to two commonly used (by a majority of the existing baselines) embedding dimensions: $512$ and $1600$. As can be seen in \cref{tab:conv4-ablation}, the network performs best with an embedding dimension of $512$ (\cfb{}). Performance is notably lower with an embedding dimension of $1600$ (\cfs{}). We hypothesize that this behavior can be attributed to the lower number of channels in the final feature map of a \cfs{} network, which is limited to $64$. 

\textbf{Effect of loss scaling factor $\beta$ on $\mathcal{L}_1$.} 
We observe that when $\beta=0$ the \cfs{} based model suffers the most as it loses $15\%$ accuracy compared to $\beta=0.7$, suggesting that training the CNN with a contrastive loss is crucial.
However, the \cfb{} model is not affected as strongly by the presence of this loss function.
Regardless, we set $\beta=0.7$ for both  models (\cfs{} and \cfb{}).

\textbf{Cross-domain robustness.} For the sake of completeness, and to further analyze the cross-domain performance of \ourmethod, in addition to \cref{tab:cdfsl_full}, we trained a \cfs{} model on CUB and tested it on tasks derived from \miniImagenet. CUB consists of $200$ classes of only birds, while \miniImagenet{} consists of $64$ classes, of which only $3$ training classes are birds.
Thus, CUB has a diminished class diversity compared to \miniImagenet. \Cref{tab:results_cub2mini} demonstrates that when training classes are diversity constrained, our method offers a better cross-domain transfer accuracy compared to the only two other competing baselines that report experimental results on this setting.


\section{Concluding Remarks}\label{sec:conclusion}
We introduced \sampclr{}, a novel contrastive pre-training method for unsupervised few-shot classification. \sampclr{} learns its representations by looking beyond single-image instances owing to a built-in self-attention message passing (\samp) module. We also propose an optimal transport (OT) based fine-tuning strategy (\opttune) which enables the creation of more representative prototypes by inducing task-awareness. Together, they construct our overall unsupervised FSL approach (coined as \ourmethod{}). 
We demonstrate that \ourmethod{} sets a new state-of-the-art for unsupervised FSL in both \miniImagenet{} and \tieredImagenet{} datasets, as well as offering competitive performance on the challenging CDFSL  benchmark \cite{guo2019new}. As future work, we are investigating the idea of incorporating memory modules in \sampclr{} pre-training to help better approximate the data distribution.  

{\small
\bibliographystyle{ieee_fullname_natbib}

\bibliography{main}
}

\end{document}